\documentclass[11pt]{article}
\usepackage{cite}
\usepackage{amsmath,amssymb,amsfonts}
\usepackage{graphicx}
\usepackage{textcomp}
\usepackage{xcolor}

\usepackage[utf8]{inputenc}
\usepackage{algpseudocode}
\usepackage{booktabs}
\usepackage{hyperref}

\def\BibTeX{{\rm B\kern-.05em{\sc i\kern-.025em b}\kern-.08em
    T\kern-.1667em\lower.7ex\hbox{E}\kern-.125emX}}

\usepackage{authblk}
\usepackage{algorithm}
\makeatletter
\newcommand{\linebreakand}{%
  \end{@IEEEauthorhalign}
  \hfill\mbox{}\par
  \mbox{}\hfill\begin{@IEEEauthorhalign}
}
\makeatother

\begin{document}

\title{Do LLMs Dream of Discrete Algorithms?}

%\author{
\author[1,2]{Claudionor N. Coelho Jr, PhD/MBA}
\author[1]{Yanen Li, PhD}
\author[1,3,4]{\\ Philip Tee}

\affil[1]{Zscaler Inc}
\affil[2]{ECE Department, Santa Clara University}
\affil[3]{Department of Informatics, University of Sussex}
\affil[4]{The Beyond Center for Fundamental Science, Arizona State University}

%\IEEEauthorblockN{Claudionor N. Coelho Jr, PhD/MBA}
%\IEEEauthorblockA{
%\textit{Zscaler Inc.} \\
%\textit{} \\
%\textit{ECE Department} \\
%\textit{Santa Clara University}}
%\and
%\IEEEauthorblockN{Yanen Li, PhD}
%\IEEEauthorblockA{\textit{Zscaler Inc.}}
%\and
%\IEEEauthorblockN{Phil Tee}
%\IEEEauthorblockA{\textit{Zscaler Inc.}}
%}

%\date{\textbf{This work has been submitted to the IEEE for possible publication. Copyright may be transferred without notice, after which this version may no longer be accessible.} \\ [0.2in] 
%\today}

\maketitle
% Print the title, author, and date

% Abstract
\begin{abstract}
Large Language Models (LLMs) have rapidly transformed the landscape of artificial intelligence, enabling natural language interfaces and dynamic orchestration of software components. However, their reliance on probabilistic inference limits their effectiveness in domains requiring strict logical reasoning, discrete decision-making, and robust interpretability. This paper investigates these limitations and proposes a neurosymbolic approach that augments LLMs with logic-based reasoning modules, particularly leveraging Prolog predicates and composable toolsets. By integrating first-order logic and explicit rule systems, our framework enables LLMs to decompose complex queries into verifiable sub-tasks, orchestrate reliable solutions, and mitigate common failure modes such as hallucination and incorrect step decomposition. We demonstrate the practical benefits of this hybrid architecture through experiments on the DABStep benchmark, showing improved precision, coverage, and system documentation in multi-step reasoning tasks. Our results indicate that combining LLMs with modular logic reasoning restores engineering rigor, enhances system reliability, and offers a scalable path toward trustworthy, interpretable AI agents across complex domains.
\end{abstract}

\section{Introduction}

Large Language Models (LLMs) have rapidly advanced the field of artificial intelligence, enabling machines to generate fluent text, answer questions, and even write code at a level that often rivals human performance~\cite{kyle:24}. Yet, a fundamental question remains: do LLMs truly possess reasoning capabilities, or are they simply leveraging statistical correlations learned from massive datasets~\cite{subbarao:25a,subbarao:25b,illusion-of-thinking}? 

This question is not just theoretical. As LLMs are starting to play an important role  in decision making systems where wrong outcomes may involve potential financial or life loss, it is imperative to characterize their limitations, so that we can make these systems more robust.  A particularly demanding challenge arises when LLMs are tasked with planning and orchestration—interpreting user intent, decomposing complex queries, and coordinating multiple software components to deliver robust solutions. These scenarios require a deeper capacity for understanding, abstraction, and structured problem-solving. While LLMs excel at tasks that rely on probabilistic inference—such as translation, summarization, or sentiment analysis—they often struggle with problems that require strict logical reasoning or adherence to formal rules~\cite{shao2025cottruereasoningjust,icaart24,shi-etal-2024-ehragent,stechly2024selfverificationlimitationslargelanguage}. Tasks like mathematical proofs, logic puzzles, or precise legal interpretations expose the limitations of current models, which may generate plausible but ultimately incorrect or inconsistent answers.

Recent research has explored augmenting LLMs with external tools, code execution, or verification mechanisms to address these gaps~\cite{nguyen2024dynasaurlargelanguageagents,smolagents,crewai,ding2024reasoning,lin2024graphenhancedlargelanguagemodels}. However, unconstrained code generation introduces new risks—such as security vulnerabilities and unpredictable behaviors—and does not fundamentally solve the challenge of robust, interpretable reasoning~\cite{cisa2024oscommand,karpathy-software-changing}.

\begin{figure}[htp]
    \centering
    \includegraphics[width=9cm]{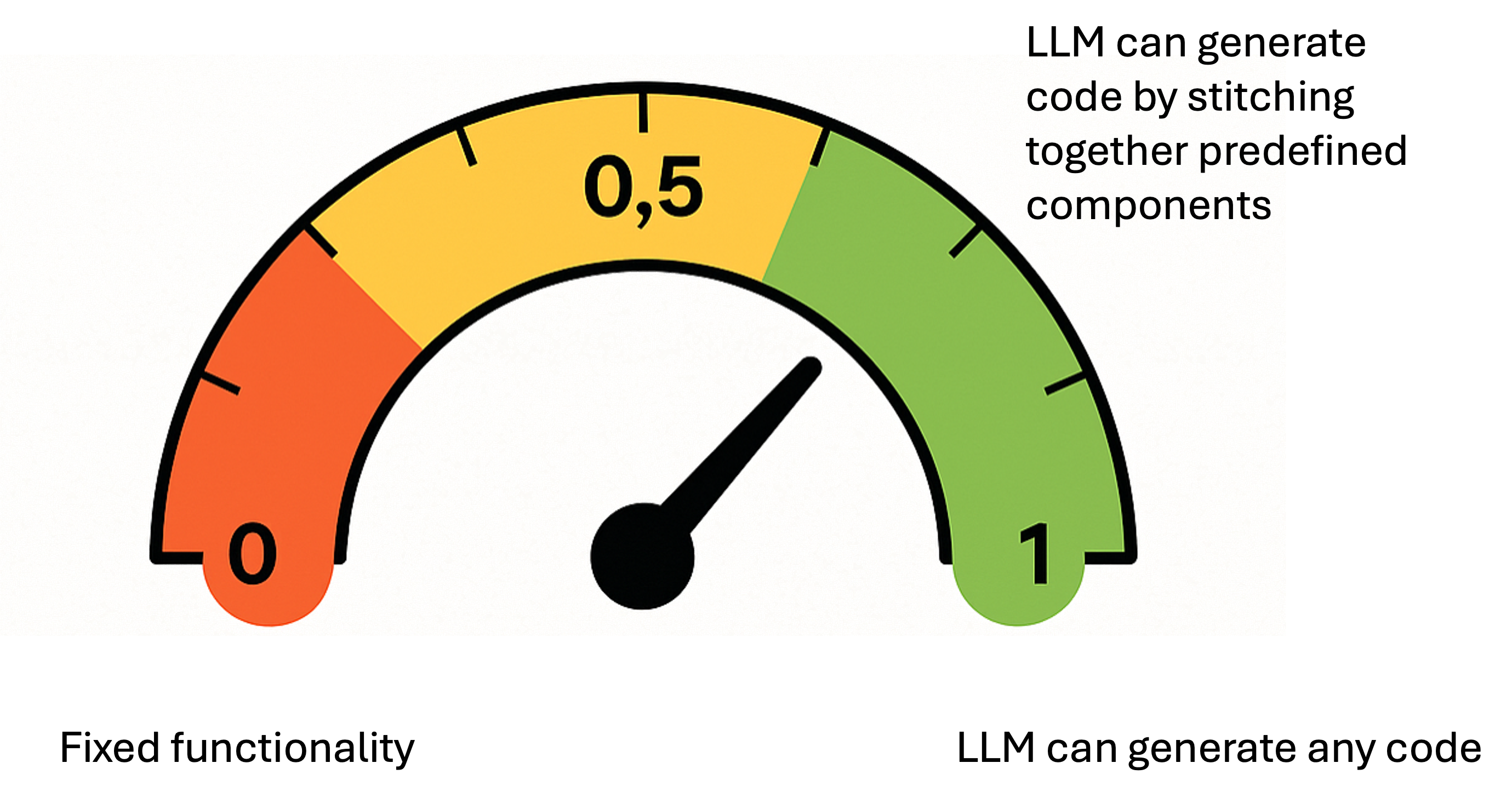}
    \caption{Trade-off between fixed functionality and complete freedom to generate any code.}
    \label{fig:speedometer}
\end{figure}

To bridge this gap, we propose a new approach: equipping LLMs with a broad set of specialized tools that can be composed to solve problems in smaller, manageable steps. Rather than aiming for unrestricted open-domain problem solving, our goal is to expand LLM capabilities across a wider range of topics by enabling them to orchestrate solutions through the intelligent composition of predefined components. This strategy, illustrated in Figure~\ref{fig:speedometer}, represents a spectrum: on one end, systems are limited to fixed functionality, replicating conventional software engineering development processes; on the other, LLMs are used generate arbitrary code. Our approach seeks a balance, empowering LLMs to generate solutions by stitching together modular, well-understood components—thus increasing reliability, interpretability, and security.

This paper is organized as follows. Section~\ref{logic} discusses the what we mean by logic, that provides a
discrete way to reasoning about the world.  Section~\ref{reasoning} discusses the limitations of LLM reasoning,
and what may happen if you mix discrete reasoning with probabilistic reasoning. Section~\ref{prolog} discusses how
we can enhance reasoning skills by adding facts, rules and a first order proof system to assist LLMs.  In Section~\ref{intelligent-agents}, we show how AI Agents can benefit from the Neurosymbolic flow. In Section~\ref{llm-new-ui}, we show that LLM-based systems with strong reasoning agents change the way we develop systems. In Section~\ref{experiments}, we show some experiments, followed by conclusions.

\section{L is for Logic}
\label{logic}

Logical reasoning~\cite{logic} is foundational to structured problem-solving in mathematics, science, and computer science. At its core, logic provides a formal framework for analyzing situations, evaluating arguments, and drawing valid conclusions based on well-defined rules and relationships. This kind of reasoning is not only essential for human cognition but also underpins many aspects of artificial intelligence, including the design of systems that can interpret, manipulate, and reason about information.

There are two primary forms of logical reasoning: deductive and inductive. Deductive reasoning starts with general premises and derives specific, guaranteed conclusions—if the premises are true, the conclusion must be true. Inductive reasoning, in contrast, involves generalizing from specific observations, leading to conclusions that are probable but not certain. While both forms are valuable, deductive reasoning is particularly important for tasks that require absolute correctness and reliability.

To formalize logical reasoning, two widely used systems are propositional logic and first-order logic (FOL). Propositional logic operates on statements that are either true or false, using logical connectives such as "and" ($\wedge$), "or" ($\vee$), and "not" ($\neg$). For example, consider the following scenario:

\begin{itemize}
    \item $p \rightarrow r$ (if $p$ then $r$)
    \item $r = \neg \text{day}$ ($r$ is true if it is not day)
    \item $\text{day}$ (it is day)
\end{itemize}

%\[
%\begin{array}{l}
%p \rightarrow r \\
%r = \neg {\tt day} \\
%{\tt day} \\
%\hline
%\neg p  
%\end{array}
%\]

From these, we can deduce using traditional logic deduction systems, such as {\sl modus tollens}, that $\neg p$ (not $p$) holds.
While propositional logic is powerful for simple truth-functional relationships, it cannot express relationships between objects or quantify over domains. Moreover, determining the validity of propositional logic statements is computationally challenging (NP-complete), which can make large-scale reasoning intractable.

First-order logic extends propositional logic by introducing quantifiers like "forall" ($\forall$) and "exists" ($\exists$), enabling reasoning over individuals, their properties, and relationships. For example, a query such as “Give me the top-5 countries with anomalous purchases” can be represented as:

\[
\begin{array}{ll}
\forall x \in {\tt Purchases} | & {\tt IsAnomalous}(x) \wedge {\tt InState} (x, s) \wedge \\ & {\tt Accumulate}(x, s, y) \wedge {\tt Sort}(y, z, 5)
\end{array}
\]

Here, $x$ ranges over purchases, and predicates like $\text{IsAnomalous}$, $\text{InState}$, $\text{Accumulate}$ and $\text{Sort}$ capture domain-specific logic. FOL is much more expressive than propositional logic, making it suitable for modeling complex domains. However, this expressiveness comes at a cost: reasoning in FOL can be undecidable, meaning that no algorithm can guarantee a solution for every possible case.

Translating user queries into logical statements, and then decomposing them into actionable steps, presents several challenges for LLMs:
\begin{enumerate}
\item Breaking down complex queries into smaller, logically coherent steps;
\item Selecting or constructing functions to perform each step;
\item Composing these steps into an overall solution that is both correct and interpretable.
\end{enumerate}

Current LLM architectures, while powerful in generating plausible text, often lack the explicit structure needed for rigorous logical reasoning and reliable problem decomposition. In the following sections, we examine why this is the case and how integrating formal logic systems with LLMs can help overcome these limitations. Our approach emphasizes the use of pre-defined, secure functions and limits arbitrary code generation to reduce security risks and improve reliability~\cite{sakana-ai-scientist,bleeping-asana-mcp}.

\section{Embeddings, Attention and Logic Reasoning}
\label{reasoning}

The development of embeddings has fundamentally changed how large language models (LLMs) represent and manipulate language~\cite{nlp}. Instead of treating words as isolated symbols, embeddings map words, phrases, and even complex concepts into high-dimensional vector spaces. This mapping allows LLMs to capture subtle semantic relationships—such as the association between "sunny" and "day"—by measuring distances or angles (for example, cosine similarity) between their respective vectors. As a result, LLMs can generate text that is contextually rich and coherent, reflecting the nuanced ways in which humans use language.

However, this statistical approach to language understanding comes with inherent limitations, especially when it comes to logical reasoning. Embeddings are exceptionally good at capturing patterns and correlations in data, but they do not encode the rigid, rule-based structures that underpin formal logic. For instance, while "sunny" and "day" may be closely related in many contexts, they are not logically equivalent. This distinction becomes critical when a task requires strict logical inference rather than probabilistic association.

The introduction of the Transformer architecture~\cite{attention} further advanced LLM capabilities by enabling models to dynamically focus on different parts of an input sequence. Through self-attention mechanisms, models can weigh the relevance of each token in context, allowing for more sophisticated pattern recognition. In practice, this means that when an LLM encounters a paragraph that can be mapped into the following logic statements:

\begin{itemize}
    \item $p \rightarrow r$ (if $p$ then $r$)
    \item $r = \neg \text{day}$ ($r$ is true if it is not day)
    \item $\text{sunny}$ (it is sunny)
\end{itemize}

\begin{figure}[htp]
    \centering
    \includegraphics[width=12cm]{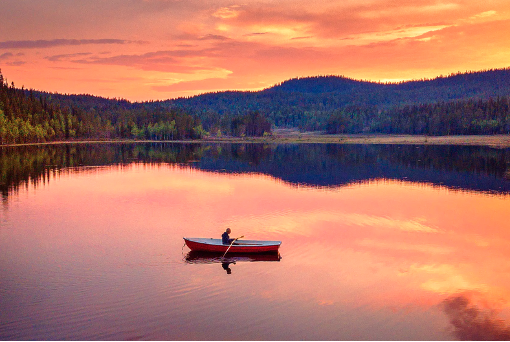}
    \caption{Midnight sun in Sweden, during the summer, according to~\cite{visit-sweden}.}
    \label{fig:midnight-sun}
\end{figure}

An LLM may use the similarity between "sunny" and "day" to infer $\neg p$, as shown in the previous deduction. Yet, this inference is based on statistical proximity in the embedding space, not on a formal logical equivalence. As illustrated in Figure~\ref{fig:midnight-sun}, such assumptions can break down in real-world scenarios—like the phenomenon of the midnight sun in Sweden—where "sunny" and "day" do not always align.   This brings a fundamental question when using LLMs for reasoning. We humans, understand clearly when to switch between probabilistic reasoning and logical reasoning. Can LLMs do the same?

This example highlights a broader challenge: LLMs are fundamentally probabilistic systems. They excel at generalizing from data and generating plausible responses, but their reasoning remains approximate. When faced with problems that require discrete, binary decisions—such as those found in mathematics or formal logic—LLMs may produce answers that are statistically likely but logically unsound. This is particularly problematic for tasks where correctness is non-negotiable, such as legal reasoning, scientific proof, or safety-critical decision-making.

Recent research~\cite{davidherron,ryu2025dividetranslatecompositionalfirstorder,zhang2024rattthoughtstructurecoherent,abhyankar2025hstarllmdrivenhybridsqltext} has sought to bridge this gap by combining the strengths of embeddings and neural architectures with the rigor of symbolic logic. Hybrid approaches introduce mechanisms for translating natural language queries into formal logical representations, decomposing complex problems into smaller steps, and verifying solutions using symbolic reasoning tools like SAT solvers or rule-based engines. These neurosymbolic systems offer a path forward, enabling LLMs to harness both the flexibility of probabilistic reasoning and the precision of formal logic.

In summary, while embeddings and attention-based algorithms have unlocked remarkable advances in language understanding and generation, they are not a substitute for true logical reasoning. To build AI systems capable of robust, reliable problem-solving across a broad range of domains, it is essential to integrate these statistical methods with explicit logical frameworks. This integration not only enhances the reasoning capabilities of LLMs but also helps ensure that their solutions are both interpretable and trustworthy.

\section{Facts, Rules and Full First Order Reasoning}
\label{prolog}

If you ask a person from the Knowledge Graph~\cite{knowledge-graph} domain what you need to complement the LLMs’ reasoning ability, he/she will say graphs or meta-annotations, i.e. vertices specifying portions of the domain expertise or knowledge connecting to ideas and concepts, and edges connecting them. In the most basic form, you may want to represent ideas that may be difficult to the LLM, either because the LLM has not been trained on them, or because you need a crisp rather than a stochastic answer in the matter. 

Dealing with discrete problems has been the center of Computer Science~\cite{history-cs}, which has existed for over 70 years, evolving from early theoretical foundations to a vast and dynamic field. Since its inception, much of its focus has been on discrete aspects of computation, dealing with structured, rule-based systems rather than continuous processes found in physical sciences. Topics such as algorithms, data structures, logic, artificial intelligence, and complexity theory form the backbone of computer science, emphasizing discrete mathematical principles. This discrete nature is evident in areas like programming languages, which operate on finite sets of instructions, and digital circuits, which rely on binary states to perform computations. While continuous paradigms like machine learning and signal processing have become increasingly influential, the fundamental role of discrete reasoning remains indispensable. Human cognition itself is often structured around discrete, categorical distinctions—logical inference, symbolic manipulation, and decision-making processes—all aligning closely with the discrete nature of computation.

We will be using throughout this article the programming language {\bf Prolog}~\cite{prolog-a, prolog-b}, although there are several other programming language alternatives that deliver the similar capabilities, such as HOL~\cite{hol}, Lean~\cite{lean-lang} or Z3~\cite{z3}, although with Z3, the representation would be at a much lower level.  

The reasons why we chose Prolog as opposed to other languages, such as the ones mentioned above, are the following:

\begin{itemize}
    \item We want to be able to represent programs as predicates without side effects, as side effects of programming language makes it harder to evaluate how information flows.
    \item We want an efficient way to represent not only the logical aspects modeling user's questions, but also a small to medium size database representing facts and rules.
    \item Even though LLMs have been trained extensively in any programming language, it has been our experience that they may struggle if the interfaces of the language or libraries change frequently (or example in different versions of C++). In this case, we may benefit from using a n more stable and older programming languages, such as Prolog and C.
    \item We wanted a language that is rich, but not too rich that will cause hallucination to become a problem, as we want to define programs as composable objects.
\end{itemize}

\subsection{Facts}

Facts express statements of a domain, such as: 

\begin{verbatim}
acquirer_country(gringotts, gb).
\end{verbatim}

This fact expresses the property that Gringrotts is an acquirer located in GB. In Prolog, variables beginning with lowercase express atoms or predicates, i.e., {\tt gringrotts} and {\tt gb} are atoms and {\tt acquirer\_country} is a predicate, expressing a relation between the atoms {\tt gringrotts} and {\tt gb}.

In general, predicates express relations that are n-ary, such as:

\begin{verbatim}
merchat_data(yoga_masters, 2, gringrotts, 7997, f)
\end{verbatim}

You can now represent several a database containing several facts:

\begin{verbatim}
acquirer_country(gringotts, gb).
acquirer_country(the_savings_and_loan_bank, us).
acquirer_country(gringbank_of_springfieldotts, us).
acquirer_country(dagoberts_vault, 'nl').
acquirer_country(dagoberts_geldpakhuis, 'nl').
acquirer_country(lehman_brothers, us).
acquirer_country(medici, it).
acquirer_country(tellsons_bank, fr).
\end{verbatim}

The reader may be asking the following questions.

\begin{itemize}
    \item Why can’t we represent these facts in a context to an LLM? In fact you can, but we are using a more compact representation.  Furthermore, prolog enables us to establish relations that are not probabilistic, such as when you specify these facts in an LLM.  When we say {\tt Gringrotts} is an acquirer located in {\tt GB}, we are not saying that Gringrotts in an acquirer located in GB with 90\% chance, but always. Second, when we state that GB is located in Europe, but it is not part of the European Union, although GB was once part of the European Union, it may have this relation probabilistically, but we are stating a precise definition of what we are defining, and that does not include GB as part of the European Union.
    \item Why can't we just use facts from an SQL database? In fact, you can. Having large tables represented in database systems will be used later on, but being able to represent part of the data or even caching important data in memory as facts pose some advantages. In addition, as we will see it later, at a minimum, you must be able to represent domains of data in Prolog to enable negation.
    \item Facts allow us to represent knowledge graphs and meta information, such as  $u \xrightarrow{\text{acquirer\_country}} v$.
\end{itemize}

\subsection{Rules}

Once we define facts, we can define rules, which are compositional rules using facts.  But to first define rules, we need to define the domain of the variables, as to perform the operation $\exists X$ or $\forall X$, we have first to define the domain of possible values of $X$.

\begin{verbatim}
acquirer(gringotts).
acquirer(the_savings_and_loan_bank).
acquirer(gringbank_of_springfieldotts).
acquirer(dagoberts_vault).
acquirer(dagoberts_geldpakhuis).
acquirer(lehman_brothers).
acquirer(medici).
acquirer(tellsons_bank).
country(gb).
country('nl').
country(us).
country(it).
country(fr).
\end{verbatim}

Once you define a domain, we can represent rules that applies to the domains.

\begin{verbatim}
acquirers_in_same_country(X,Y):-
    acquirer(X), acquirer(Y), country(Z),
    X \= Y,
    acquirer_country(X,Z), acquirer_country(Y,Z).
\end{verbatim}

This rule states that $X$ and $Y$ are acquirers in the same country
if there is a country $Z$ (in first order logic, $\exists Z \in {\tt Domain}$) such that $X$ is an acquirer $Z$ and $Y$ is an acquirer in $Z$. Using {\tt acquirer(X)} and {\tt acquirer(Y)} ensures that in the event the user did not specify $X$ or $Y$, e.g., when they are used as output variables, we will be doing $\forall X \in {\tt Acquirer\_Domain}$ and $\forall Y \in {\tt Acquirer\_Domain}$, respectively. In the definition of \texttt{acquirers\allowbreak \_in\allowbreak \_same\allowbreak \_country\allowbreak (X,Y)}, \texttt{\string\=} means not equal.

Prolog, having first order representation of symbolic programs, allows you to specify more complex relations, such as negations.

\begin{verbatim}
not_in_same_country(X, Y) :- 
   acquirer(X), acquirer(Y), country(Z),
   X \= Y,
   acquirer_country(X, Z),
   \+ acquirer_country(Y, Z).
\end{verbatim}

In this case, \texttt{\string\+ acquirer\_country(X, Z)} means {\tt acquirer\_country(X, Z)} fails, which may be considered a form of negation.  Being able to express negation is what differentiates Prolog from traversals in knowledge graphs, and negation makes the language a first-order logic proof system.

In this example, if you query the database for \texttt{not\allowbreak \_in\allowbreak \_same\allowbreak \_country\allowbreak (lehman\_\allowbreak brothers, Y)}, we get as answer the following, using SWIPL Prolog interpreter~\cite{swipl}.

\begin{verbatim}
?- not_in_same_country(lehman_brothers, Y).
Y = gringotts ;
Y = dagoberts_vault ;
Y = dagoberts_geldpakhuis ;
Y = medici ;
Y = tellsons_bank.
\end{verbatim}

You can see now the advantages of representing facts and rules as opposed to representing just knowledge graphs. Whereas facts are equivalent to knowledge graphs, rules represent paths, but negation adds another level of complexity that is very hard to represent in pure knowledge graphs.

One important fact here is that expressing domains for very large databases may be infeasible. We have to decide on the feasibility of expressing negation and quantifiers for variables to be considered inputs or outputs, or just only accepting instantiated variables as inputs (in case we cannot represent the full domain of variables).

The reader should note that we are not saying that probabilistic search is bad. If you have a graph, or a database table with transactions, and you want to know which pair of connected nodes or which translation may be faulty with high probability, you are referring to a probabilistic search. On the other hand, if you want to know if a node can reach a destination node of a graph, you do not care if the probability is 1 or 99\%, you want to know if there is a path to that destination. That's a binary answer (yes/no), and that’s where a language representing facts, rules and proof system such as Prolog sits in.

Prolog is a rich language, and we will mention a few operators that are worth mentioning (non-exhaustive):

\begin{itemize}
    \item 	$P1(X),!,P2(X,Y)$: finds an assignment for variables in $P1$ (e.g. $X$ in our example), then for all solutions $P2$ finds a solution (e.g. $Y$ in our example), but does not search for an alternative $X$ in $P1$. It can be thought of as converting a $\forall X | P1(X)$ into $\exists X | P1(X)$, i.e., it finds the first match to $P1(X)$ and it never re-evaluates it.  

    In our example, it becomes, $\exists X | P1(X) \wedge \forall Y | P2(X,Y)$ with the caveat that once a solution for $P1(X)$ is satisfied, we will only look for values in the domain of  Y that satisfies $P2(X,Y)$.

    For example, the operator \texttt{\string\+ P} can be defined as:

    \begin{verbatim}
    negate(P) :- call(P), !, fail.
    negate(_).
    \end{verbatim}

    In other words, if calling {\tt P} succeeds, {\tt negate(P)} fails, otherwise it succeeds.
    \item 	{\tt findall(Variable, Goal, Result)}: equivalent to {\tt Result} = \\ $\left[ \forall {\tt Variable} \in {\tt Goal(Variable)} \right]$.
    \item 	{\tt maplist(Function, List, Result)}: equivalent to {\tt Result} = \\ $\left[ \forall X \in {\tt List} | {\tt Function(X)} \right]$.
    \item \texttt{\string Term =.. [Func|Args], call(Term)}: allows you to do partial matching or dynamic matching of functions.
    \item \texttt{\string current\_op(Precedence, Type, Name)}: given an operator, what's its precedence and type (the three variables can be inputs or outputs).
    \item \texttt{\string current\_predicate(Name/Arity)}: this predicate searches for name of the function and its arity.
\end{itemize}

These last functions are useful to create dynamic infrastructure for facts, rules, and proofs, and searching for their information.  The reader should know that Prolog is a much richer function with native support for lists, with \texttt{[1,2,3]} representing a list of three elements, and \texttt{[X|Xs]} representing a list where \texttt{X} is the first element of the list, and \texttt{Xs} representing the remaining elements of the list.

\section{Intelligent Agents}
\label{intelligent-agents}

\begin{figure}[htpb]
    \centering
    \includegraphics[width=13cm]{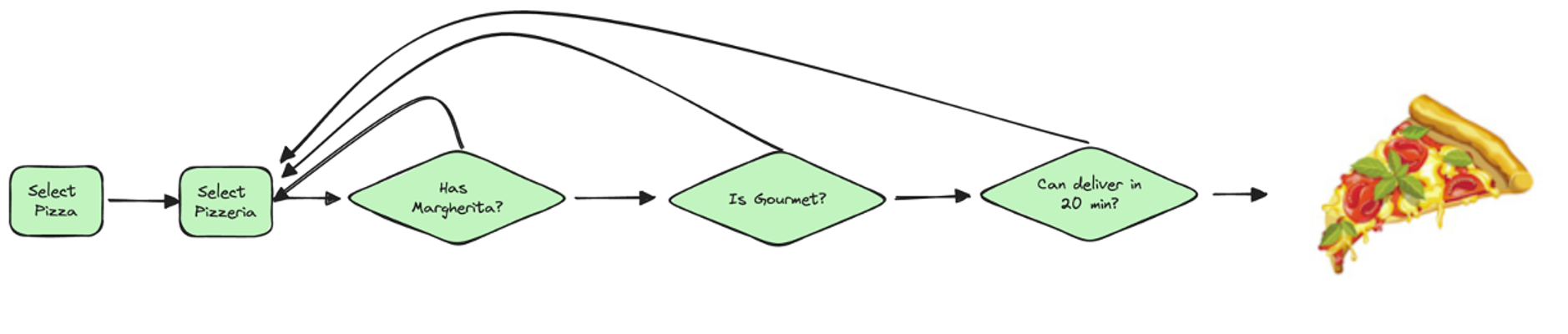}
    \caption[User Story]{User Story to Order Pizza for a Food Delivery App} % \footnotemark}
    \label{fig:order-pizza}
\end{figure}

We will be discussing in this section how we can transition from a use-case or MVC software development to intelligent agents using LLMs as interfaces. In the following example, we want to examine the complexity of adding the user story of ordering a Margherita gourmet pizza in 20 minutes to a food app, as 
an optimization to the the flow presented in Figure~\ref{fig:order-pizza}.

We have to assume that to implement this use case, we need access to the following data sources and algorithms:

\begin{enumerate}
    \item Restaurant database that can be searched by location and by type of food.
    \item Menu database, where user can search for types of food served by the restaurant.
    \item Algorithm that computes the delivery time from the restaurant to your location.
\end{enumerate}

The reader should notice that this use case implements a {\bf single} type of user interaction, and if we decide to modify the interaction, we will need to change the user story, or create another implementation that accommodates a different user story.

In general, given as sample of user's queries about a system, we want to be able to generate complete systems that generalize and are able to answer a very large number of questions.  In~\cite{coelho}, we addressed the problem industry is facing right now that with LLMs, and more accurately, with AI Agents, specifications for software systems are appearing as a set of questions the system needs to answer. We addressed this problem by using LLMs to enlarge the few questions people provide to hundreds of questions of similar nature in an environment, and later generating maps to {\bf algorithms}, {\bf data sources} and {\bf UI/UX interactions or visualizations}.

\subsection{AI Agents}

An AI Agent~\cite{xi2023risepotentiallargelanguage,praveen-akkiraju,nvidia2023} encompasses a system that employs an LLM to process and reason about a specific domain. To generate specific answers (often related to the domain), the AI Agent leverages auxiliary systems in conjunction with the LLM. These auxiliary systems support the agent in comprehending the domain and facilitating the creation of accurate responses.

AI Agents consist of four major components. The \emph{agent core}  
forms the central component and is responsible for orchestrating the agent's overall functionality. The \emph{memory} module enables the agent to store and retrieve relevant information, enhancing its ability to retain context and make informed decisions. The \emph{planner} component guides the agent's actions by formulating a strategic course of actions based on the given problem or task. Finally, the set of \emph{tools} encompasses various external components and resources that assist the agent in performing specific tasks or functions within the defined domain, such as algorithms and data queries. These components collaboratively enable AI Agents to effectively process information, reason, adapt to query changes and generate responses in a manner aligned with their designated purpose.

\begin{figure}[htp]
    \centering
    \includegraphics[width=7cm]{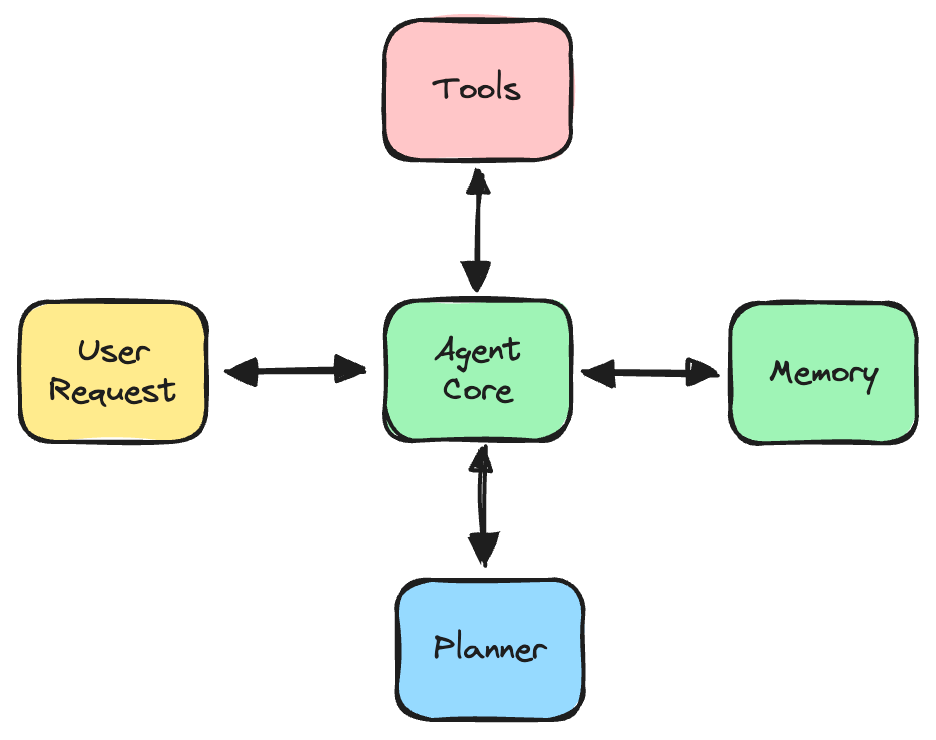}
    \caption{AI Agent from~\cite{nvidia2023}}
    \label{fig:ai-agent}
\end{figure}

\subsubsection{Agent Core}

The agent core is a crucial component within an AI Agent that plays a central role in orchestrating the agent's overall functionality. It receives a query from the user. Consequently, it manages the decision-making processes, communication, and coordination of various modules and subsystems within the agent. Finally, it aggregates the information and generates a response. %The primary function of the agent core is to facilitate the seamless operation of the AI Agent and ensure efficient interaction with the environment or the tasks at hand.

%The agent core acts as the interface between the AI Agent and the user. It receives a query from the user, coordinates exchange and processing of information with the other components and generates a response.
%from the environment or external systems, processes the information, and generates appropriate actions or responses. 
%This involves employing various algorithms, heuristics, or decision-making mechanisms to analyze the received data and determine the best course of action. 
%The core also handles the coordination of different modules and subsystems within the AI Agent, ensuring that they work in harmony to achieve the agent's objectives.

The agent core is also responsible for managing the agent's internal state. It maintains a representation of the agent's assets and internal state, allowing it to reason, plan, and adapt its behavior accordingly. The core oversees the update and retrieval of information from the agent's memory, enabling it to access relevant knowledge and contextual information during decision-making processes.

%Overall, the agent core acts as the brain of an AI Agent, providing the necessary intelligence, coordination, and control to effectively interact with the environment and perform tasks within the defined domain. It governs the decision-making, communication, and coordination processes, ensuring the agent operates optimally and achieves its objectives.

\subsubsection{Memory}

The memory module within an AI Agent encompasses two important aspects: historical memory~\cite{wu2025humanmemoryaimemory} and contextual memory, usually represented as a knowledge-graph~\cite{zuo2025kg4diagnosishierarchicalmultiagentllm}. 

{\bf Historical memory} serves as a repository for past interactions and experiences of the AI Agent. It stores a record of previous inputs, outputs, and the outcomes of actions taken by the agent. This historical data is valuable as it enables the agent to learn from past interactions and avoid repeating mistakes. Through the historical memory the agent gains insights about effective strategies, successful outcomes/patterns enabling an informed decision making process. Historical memory is usually split as short term memory and long term memory. Short term memory is usually retrieved from temporal most recent interactions, whereas long term memory is usually stored and retrieved through closeness to the topic being discussed.

{\bf Contextual memory} maintains an understanding of the current and coherent context, and the world as we perceive, in the context of the objective on the AI Agent.
It stores relevant context that provides the necessary background for the agent to interpret and respond appropriately to the present state. This can include information about the environment, the user's preferences or intentions, and any other contextual factors that influence the agent's behavior. Contextual memory allows the agent to adapt its action and responses to specific circumstances, thereby enhancing its ability to interact intelligently with changing environments.

Together, historical and contextual memories allow the AI agent to combine past experiences and current context for an efficient decision making process.

\subsubsection{Planner}

The planner component within an AI Agent plays a crucial role in guiding the agent's actions and formulating a strategic course of action based on the given problem or task. It is responsible for generating a sequence of steps or actions that lead the agent towards achieving its objectives. The planner analyzes the current state of the environment, along with any available information or constraints, to determine the most effective sequence of actions (by calling functions or making queries to databases) to achieve the desired outcome. It also takes into account other factors such as goals, resources, rules, and dependencies to generate a plan that optimizes the agent's decision-making process.

When the user asks the question \textit{I want to order a Margherita pizza in 20 min}, the planner needs to consolidate information about user's temporal requests, context (e.g. the location where the user is located), tools and data sources available, and finally develop a detailed plan of action. In later sections, this is where we will focus our attention, i.e. on how to assist AI Agents with logical reasoning in order to be able to answer complex questions, or deal with complex scenarios. 

\subsubsection{Tools}

In an AI Agent, the set of tools encompasses various resources and functionalities that assist in performing specific tasks or functions within the defined domain. Here is a non-exhaustive list of possible tools that can be utilized in an AI Agent:

\begin{itemize}
\item {\it RAG (Retrieval-Augmented Generation~\cite{lewis2021retrievalaugmentedgenerationknowledgeintensivenlp})} ---  Combines retrieval-based methods with generative language models. It enables the agent to retrieve relevant information from a knowledge base and use it to generate coherent and contextually appropriate responses.  Common data sources for RAG include Question-Answer databases, documentation and web pages.

\item {\it Structured data access} --- Connect to databases and allow the AI Agent to dynamically access and retrieve information from structured external data sources. 

\item {\it Unstructured data access} --- Connect to a plethora of multi-modal sources of information (text, pictures, video streams, PDF, power-point presentations, spreadsheets, to name a few), some of which can be indexed by RAG systems, some of which may be accessed and retrieved by some other means, such as search.

\item {\it Machine learning frameworks} --- Provide tools and algorithms for training and deploying machine learning models. These frameworks enable the agent to use various machine learning techniques, including supervised learning, unsupervised learning, or reinforcement learning, to enhance its capabilities.

\item {\it Visualization tools} --- Help represent and interpret data or model outputs in a visual format. These tools can help the agent understand complex patterns, relationships, or trends in the data, helping in decision-making and analysis.

\item {\it Simulation environments} --- Provide a controlled virtual environment where the AI Agent can interact and learn without impacting the real world. These tools allow the agent to practice and refine its skills, test different strategies, and evaluate the potential outcomes of its actions.

\item {\it Monitoring and logging frameworks} --- Monitoring and logging frameworks facilitate the tracking and recording of agent activities, performance metrics, or system events. These tools assist in evaluating the agent's behavior, identifying potential issues or anomalies, and supporting debugging and analysis.

\item {\it Data preprocessing tools} --- Help in cleaning, transforming, and preparing raw data before feeding it into the AI Agent. These tools may include techniques for data cleaning, normalization, feature selection, or dimensionality reduction, ensuring the quality and relevance of data used by the agent.
\end{itemize}

These tools enhance LLM capabilities by providing it with specialized functionalities for specific domains that would be outside the domain of LLMs.  Any system using LLMs as interfaces to user interactions, such as the Copilot in~\cite{Srivastava2024} will incorporate these components into the implementation.

In this paper, we will use the classification of~\cite{coelho} to label tools as data sources, visualization artifacts and algorithms.

\section{LLM is the New UI/UX}
\label{llm-new-ui}

With advent of LLMs and intelligent agents, it has become natural for people to desire LLMs to being able to answer questions like the following question.

\begin{verbatim}
I want to order a gourmet Margherita pizza in 20 minutes.
\end{verbatim}

When used as the front end to human interaction, LLM-based user interfaces enable systems to dynamically change queries and algorithm executions, depending on user's intent, and to route the correct visualization to the user. As mentioned in Gartner's keynote~\cite{gartner-keynote}, \textit{\string the most obvious value of of LLMs is to put the human in the center of the interactions of systems, and this provides a radical simplification of the use of such systems}.  This simplification is achieved by being able analyze information, and the adaptation comes from being able to change the flow of control and access to data to accommodate what the user requests. In other words, LLMs can write their own code and access data sources in patterns we may not have thought before.

The problem with that this previous statement is that it carries a great deal of uncertainty that software engineers are required to answer before they can implement the system. In user story development, for example, as follow-up questions one would need to document in the development plan include:

\begin{itemize}
    \item Which data sources should we connect to?
    \item Which algorithms do we need to invoke to solve this request?
    \item Which interfaces are required to implement this user story?
    \item How do we ensure levels or privacy and security, given the user's query may trigger a dynamic program and data access patterns ?
    \item Which other questions do we want to be able to solve?
\end{itemize}

Together with that simple question, and in order to make systems more generalizable~\cite{Gibbons2023}, we need to understand the scope of such systems, and possibly being able to answer a broader set of questions, such as:

\begin{enumerate}
    \item Can this restaurant deliver food in 20 min?
    \item Give me the list of all restaurants that deliver gourmet pizza in 20 min.
    \item Give me the 20 top evaluated restaurants that can deliver gourmet pizza in 20 minutes.
\end{enumerate}

The reader can easily see that the first question requires just a simple yes/no answer. The second question requires a summarization or visualization agent to provide the answer. The third query will require getting data from possibly an additional table from the backend.  Without fully understanding the scope of the system, and understanding what are the data sources, algorithms and visualizations that are needed, generalizing user's intent can easily become an almost impossible task.

\begin{figure}[htp]
    \centering
    \includegraphics[width=12cm]{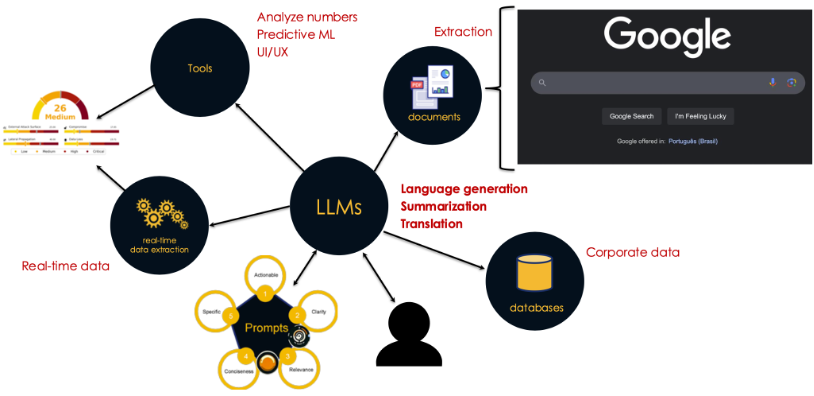}
    \caption{AI Agent Representation From Interaction View}
    \label{fig:more-complete-ai-agent}
\end{figure}

Because of the complexity of such systems, a better representation of an AI Agent interaction with different types of tools and data sources is given in~\ref{fig:more-complete-ai-agent}

\section{The Role of Prolog to Assist Planner Activities }

In order to discuss the role of logic reasoning in assisting planning activities, we have first to understand what may go wrong when a user asks a question.

\begin{itemize}
    \item First, and foremost, the LLM may hallucinate in generating the answer, as such systems are based on probabilistic generation of text~\cite{Huang_2025}.

    \item Second, if the AI Agent is searching for information using RAG, the search mechanism may not be enough to retrieve the answer to the user~\cite{fan2024surveyragmeetingllms}. In its simplest forms, RAG utilizes a proximity metric that may not align well with the user's intent. 

    \item Third, if the planner decided on the wrong set of steps to execute answer a user's question, we may end up executing the wrong query from the user. This is the problem we want to attack by adding logic reasoning~\cite{subbarao:25a,subbarao:25b}. 
\end{itemize}

It is important to note that considering symbolic and neural techniques has been proposed in the past, such as in~\cite{lamb,log2ns,pan2023logiclmempoweringlargelanguage}, and the use of Prolog in conjunction with LLMs have been proposed in several recent works~\cite{prolog-llm-era,tan2024thoughtlikeproenhancingreasoninglarge,borazjanizadeh2024reliablereasoningnaturallanguage,cot-prolog,Vakharia_2024,yang2024arithmeticreasoningllmprolog,enhancing-llm-reasoning}.

One of the major roles here is played by the planner stage, as it is responsible for outlining the steps to solve the query from the user. As we mentioned earlier, we need a solution that is able to:

\begin{itemize}
\item Define the steps as a sequence of positive and negative predicates (remember, Prolog has negation of predicates).
\item Utilize a large number of predicates that are able to cover technical questions from the users.
\item Create compositions of steps, such as and-ing or or-ing predicates.
\item Either train an LLM to understand the underlying language to specify these steps, or use a language that has already been trained in the LLM.
\item Have a very low cost of establishing ground truth, and even fix small mistakes from the LLM (if we use a programming language such as Prolog, ground truth can be established by compiling the code, and a parse tree can understand how good or bad the solution is).
\end{itemize}

Our approach here is to exercise what LLMs know best, i.e. how to process grammars, as LLMs have excelled in being able to generate text following grammatical rules.

\begin{figure}[htp]
    \centering
    \includegraphics[width=12cm]{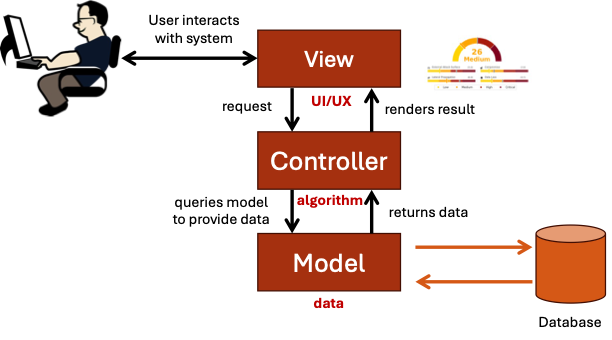}
    \caption{Model-View-Controller system}
    \label{fig:mvc}
\end{figure}

We will follow the steps defined in~\cite{coelho} to create the tools we need to solve a given set of problems, i.e., starting from a set of seed questions from the user, we enlarge these questions and attempt to generate a list representative questions that correspond to the problem to be solved, generate a set of MVC components (model-view-controller)~\cite{voorhees2021guide}, as the MVC framework nicely represents the different types of components that exists between interactions of users to systems, as shown in Figure~\ref{fig:mvc}.  Finally, from the MVC components, we identify the gaps of our implementation, and outline which functions need to be implemented.

As we want to cast this problem as a grammar problem with strong typing system, we need first to define the basic types we will use in this system, as seen below.

\begin{itemize}
    \item \texttt{StrOrList: Union[str,List[StrOrList]]}: meta type to be used by \texttt{Filter}.
    \item \texttt{TableName: str}: table according to the list of tables the system can access.
    \item \texttt{Filter: StrOrList}: filter expressions, e.g. \texttt{[ and, [issuing\_country, 'GB'], [ip\_country, 'FR']]}.
    \item \texttt{Projection: List[str]}: list of names from the table \texttt{TableName} that we will return to the user, according to the field names in the database.
    \item \texttt{Header: List[str]}: list of field names from \texttt{TableName} or \texttt{Projection} if the projected list is not empty.
    \item \texttt{Data: List[List[str]]}: list of data fields, where the size of the inner list is the same as the \texttt{Header}, and each element of the inner list of \texttt{Data} corresponds a value of each field of \texttt{Header}.    
\end{itemize}

Once we define data types used by the system, our objective is to define predicates that manipulate data.  We provide at minimum functions that can be called to retrieve data and filter data in the system.  Some examples of functions that we can use are presented in Table~\ref{table:mvc}.  This first set of functions correspond to functions available in SQL language (for now, we will keep these functions separate, although the user may see similarity between them and SQL query artifacts), as bundling them together depends on how we pay for the access to the data - e.g. do we pay per query or per byte transmitted?  In addition, we may have certain domains that can be easily expressed as facts in Prolog, if the data sizes are not too large.  The second set of functions correspond to algorithms, such as traditional predictive ML algorithms, complex arithmetic functions, etc. The third set of functions correspond to UI/UX artifacts, such as visualization of maps.

It is clear that we may still need to comment the functions defined above, explaining which fields are expected or are generated. For example, an anomaly detection algorithm may require additional information, such that notifying the Planner that an amount paid by the customer and the state
where the purchase were made are required input fields for the anomaly detection algorithm to work. Sometimes, even small examples of usage of these algorithms may be required.

The user may be asking the question, 1) how many functions is enough, 2) whether it is possible to dynamically instantiate functions in this infrastructure.

\begin{table}
\caption{List of functions available for the logic reasoner.}
\vspace*{4pt}
\label{table:mvc}
\small
    \centering
    \begin{tabular}{|l|}
    \hline
    \textbf{Function} \\ \hline 
    \texttt{query\_data(-TableName, -Filter, -Projection, +[Header | Data])} \\
    \texttt{filter(-[Header | Data], -Filter, +[Header1 | Data]} \\
    \texttt{project(-[Header | Data], -Projection, +[Projection | Data])} \\
    \texttt{count(-[Header | Data], +[Header | Data])} \\
    \hline
    \texttt{anomaly(-[Header | Data], +[[is\_anomaly | Header] | Data]} \\
    \hline
    \texttt{visualize\_map(-[Header | Data])} \\
    \hline
    \end{tabular}
\end{table}

To answer the first question, we will refer the user to~\cite{coelho}. In this paper, we have shown that software engineering has changed from a structured design process to giving a sample of questions by product managers when using LLMs as interfaces. We argued that we need in this case to extract specification by following the steps below.

\begin{enumerate}
    \item Expand the list of questions from the users using the original sample questions as seed.
    \item Extract the model-view-components from the questions, always reusing the previous components whenever possible
\end{enumerate}

The full algorithm proposed in~\cite{coelho} is described in Algorithm~\ref{fig:algorithm}.  It is worth noting that as LLMs can hallucinate, and since extraction of the specification is executed in the beginning of the project, we recommended manual review of all enlarged questions and MVC components.

\begin{algorithm}
\begin{algorithmic}
\Require List of questions $Q$
\State ${\it AllTasks} \gets \emptyset$
\For{$q \in Q$}
\State Generate $N$ related questions $Q_q$ from $q$.
\State $T_q \gets {\it Planner}($
\State $\qquad \{ q \} \cup Q_q, {\it current\_tools}=AllTasks, $
\State $\qquad minimize=True)$
\For{$t \in T_q$}
\State $t['task'] \gets {\it model}| {\it view} | {\it controller}$
\EndFor
\State Manually validate the set $T_q$.
\State ${\it AllTasks} \gets {\it AllTasks} \cup T_q$
\EndFor
\State Manually validate final set of {\it AllTasks}
\end{algorithmic}
\caption{Algorithm to extract MVC components from enlarged number of questions}
\label{fig:algorithm}
\end{algorithm}

\subsection{Planning by Stitching Together Logical Components}

Now we can list the general structure of the prompt for the planner to generate a planner strategy in Prolog.

\begin{verbatim}
You will interpret the user's query and generate a prolog code
that best matches the user's intent to the prolog code. In 
doing so, you will evaluate the prolog code accoding to the 
`evaluation` metric, and you will try to maximize the 
`evaluation` metric.

<query>
user query
</query>

Previous interactions with the user are as follows:

<history>
short term, long term interactions.
</history>

You base use all functions available in the standard prolog 
language, with the addition of the following list of prolog 
assert database and foreign functions.

<assert>
list of prolog database
</assert>

<foreign-functions>
list of foreign-functions
</foreign-functions>

Data accesses must following the following schema.

<database-schema-description>
Definition of database schema in English
</database-schema-description>

You must use the following examples to guide your reasoning:

<examples>
list Chain-of-Thought examples
</examples>

Your response must be in the following JSON format:

```json
{
    "explanation”: str,
    "gaps”: List[str],
    "findings”: List[str],
    "plan”: List[str],
    ”action”: List[str],
    ”result”: str,
    ”evaluation”: float,
}
```

explanation of each field, asking the LLM to align the 
action list to the CoT examples, and on how to increase 
and decrease the evaluation by specific quantifiable amounts.

In the evaluation, use the following guidelines:
- start with an evaluation score of 1.0 after generating
the code.
- if you utilize an input to a foreign function before 
instantiating its value, set evaluation score to 0.0.
- if you use an LLM written assert predicate, reduce the 
evaluation score by 0.2.
- if you call native Prolog functions such as sort, findall 
or aggregate with objects containing both the header and the 
data fields, reduce the evaluation score by 0.4.
...

Remember, ...

Now generate the JSON file, maximizing the evaluation score
as instructed.
\end{verbatim}

In this prompt, the list of \texttt{assert} functions corresponds to predicates written in Prolog, and \texttt{foreign-functions} corresponds to predicates written in other programming languages.

The idea of the evaluation metric is that we ask the LLM to self reflect on its reasoning steps by giving rewards and penalties.  
It is worth noting that in~\cite{zhao2025learningreasonexternalrewards}, a similar
approach has been proposed to self-generate evaluation metrics that are fed into GRPO fine-tuning loop.

To answer the second question, we have to consider the three scenarios:

\begin{enumerate}
    \item In the past~\cite{voorhees2021guide}, people have resorted to the specification of the user stories determine how the whole interaction with the user is driven, such as in the case of the Pizza ordering App, as presented in Figure~\ref{fig:order-pizza}.
    \item In~\cite{smolagents}, we are able to dynamically generate, and keep trying until code passes. As advantage, it connects directly to Python infrastructure. However, because the Python language is extremely complex (e.g. users have to specify which libraries they intend to use).
    \item We provide a large number of small components, and expect the LLM to "compose" a solution by stitching together small portions to comprise a full solution. As advantage, this allows one to have better control over hallucination and security, and even use smaller or simpler models to generate the code. On the negative side, our ability to answer very complex questions reside on the initial steps to explore the solution space. 
\end{enumerate}

We consider our approach to be a meet in the middle from a completely fixed functionality specification, such as the ones presented in traditional software engineering strategies, and the completely loose application to let the LLM freely try different implementations until one passes, which may lead to answering open-domain questions. 

\section{Dataset Overview}
\label{experiments}

\subsection{DABStep Benchmark Overview}

To develop our approach, we leveraged the \textbf{DABStep Benchmark}\footnote{\url{https://huggingface.co/blog/dabstep}}, a dataset specifically designed to assess the multi-step reasoning capabilities of AI agents. DABStep (Data Agent Benchmark for Stepwise Evaluation of Planning) comprises over 450 real-world, multi-hop reasoning tasks focused on domains such as financial transactions, risk analytics, merchant behavior, and operational metrics.

Each task includes structured data (in CSV or JSON format), optional unstructured context (such as documentation or manuals), and a natural language question that must be answered with high factual precision. The benchmark is engineered to challenge agents on planning, data selection, transformation, and reasoning, and it uses an exact-match evaluation metric that does not tolerate partial correctness or hallucination.

Table~\ref{tab:dataset-files} shows a snapshot of the released data files that comprise the DABStep benchmark.

\begin{table}[ht]
\centering
\caption{Core data files included in the DABStep benchmark}
\label{tab:dataset-files}
\tiny
\begin{tabular}{@{}ll@{}}
\toprule
\textbf{File Name} & \textbf{Description} \\ \midrule
\texttt{payments.csv} & 138k anonymized transactions with fraud/risk signals \\
\texttt{payments-readme.md} & Documentation for interpreting the payments dataset \\
\texttt{acquirer\_countries.csv} & List of acquiring banks and their corresponding countries \\
\texttt{fees.json} & Dataset of 1000 structured fee schemes from financial networks \\
\texttt{merchant\_category\_codes.csv} & Lookup table of Merchant Category Codes (MCCs) \\
\texttt{merchant\_data.json} & JSON file describing merchant identities and properties \\
\texttt{manual.md} & Condensed business handbook used for domain rules and constraints \\
\bottomrule
\end{tabular}
\end{table}

The dataset tables are summarized as follows.

\begin{itemize}
    \item acquirer\_countries.csv
    \begin{itemize}
        \item \texttt{acquirer}: string. acquirer name
        \item \texttt{country\_code}: The location (country) of the acquiring bank (Categorical) - SE, NL, LU, IT, BE, FR, GR, ES.
    \end{itemize}
\end{itemize}

\begin{itemize}
    \item fees.json
    \begin{itemize}
        \item \texttt{ID}: identifier of the fee rule within the rule fee dataset
        \item \texttt{card\_scheme}: string type. name of the card scheme or network that the fee applies to
        \item \texttt{account\_type}: list type. list of account types according to the categorization `Account Type` in this manual
        \item \texttt{capture\_delay}: string type. rule that specifies the number of days in which the capture from authorization to settlement needs to happen. Possible values are '3-5' (between 3 and 5 days), '>5' (more than 5 days is possible), '<3' (before 3 days), 'immediate', or 'manual'. The faster the capture to settlement happens, the more expensive it is.
        \item \texttt{monthly\_fraud\_level}: string type. rule that specifies the fraud levels measured as ratio between monthly total volume and monthly volume notified as fraud. For example '7.7%-8.3%' means that the ratio should be between 7.7 and 8.3 percent. Generally, the payment processors will become more expensive as fraud rate increases.
        \item \texttt{monthly\_volume}: string type. rule that specifies the monthly total volume of the merchant. '100k-1m' is between 100.000 (100k) and 1.000.000 (1m). All volumes are specified in euros. Normally merchants with higher volume are able to get cheaper fees from payments processors.
        \item \texttt{merchant\_category\_code}: list type. integer that specifies the possible merchant category codes, according to the categorization found in this manual in the section `Merchant Category Code`. eg, `062, 8011, 8021`.
        \item \texttt{is\_credit}: bool. True if the rule applies for credit transactions. Typically credit transactions are more expensive (higher fee).
        \item \texttt{aci}: list type. string that specifies an array of possible Authorization Characteristics Indicator (ACI) according to the categorization specified in this manual in the section `Authorization Characteristics Indicator`.
        \item \texttt{fixed\_amount}: float. Fixed amount of the fee in euros per transaction, for the given rule.
        \item \texttt{rate}: integer. Variable rate to be especified to be multiplied by the transaction value and divided by 10000.
        \item \texttt{intracountry}: bool. True if the transaction is domestic, defined by the fact that the issuer country and the acquiring country are the same. False are for international transactions where the issuer country and acquirer country are different and typically are more expensive.
    \end{itemize}
\end{itemize}

\begin{itemize}
    \item merchant\_category\_codes.csv
    \begin{itemize}
        \item \texttt{mcc}: integer that specifies the possible merchant category code.
        \item \texttt{description}: textual description of merchant category code.
    \end{itemize}
\end{itemize}
\begin{itemize}
    \item merchant\_data.json
    \begin{itemize}
        \item \texttt{merchant}: Merchant name (Categorical), eg Starbucks or Netflix.
        \item \texttt{capture\_delay}: string type. rule that specifies the number of days in which the capture from authorization to settlement needs to happen. Possible values are '3-5' (between 3 and 5 days), '>5' (more than 5 days is possible), '<3' (before 3 days), 'immediate', or 'manual'. The faster the capture to settlement happens, the more expensive it is.
        \item \texttt{acquirer}: list of string. acquirer names allowed for merchant. eg, `"dagoberts\_geldpakhuis", "bank\_of\_springfield"`.
        \item \texttt{merchant\_category\_code}: list type. integer that specifies the possible merchant category codes, according to the categorization found in this manual in the section `Merchant Category Code`. eg, `8062, 8011, 8021`.
        \item \texttt{account\_type}: string, one of R,D,H,F,S,O, where R = Enterprise - Retail, D = Enterprise - Digital, H = Enterprise - Hospitality, F = Platform - Franchise, S = Platform - SaaS, and O = Other.
    \end{itemize}
\end{itemize}

\begin{itemize}
    \item payments.csv
    \begin{itemize}
        \item \texttt{psp\_reference}: Unique payment identifier (ID).
        \item \texttt{merchant}: Merchant name (Categorical), eg Starbucks or Netflix.
        \item \texttt{card\_scheme}: Card Scheme used (Categorical) - MasterCard, Visa, Amex, Other.
        \item \texttt{year}: Payment initiation year (Numeric).
        \item \texttt{hour\_of\_day}: Payment initiation hour (Numeric).
        \item \texttt{minute\_of\_hour}: Payment initiation minute (Numeric).
        \item \texttt{day\_of\_year}: Day of the year of payment initiation (Numeric).
        \item \texttt{is\_credit}: Credit or Debit card indicator (Categorical).
        \item \texttt{eur\_amount}: Payment amount in euro (Numeric).
        \item \texttt{ip\_country}: The country the shopper was in at time of transaction (determined by IP address) (Categorical) - SE, NL, LU, IT, BE, FR, GR, ES.
        \item \texttt{issuing\_country}: Card-issuing country (Categorical) - SE, NL, LU, IT, BE, FR, GR, ES.
        \item \texttt{device\_type}: Device type used (Categorical) - Windows, Linux, MacOS, iOS, Android, Other.
        \item \texttt{ip\_address}: Hashed shopper's IP (ID).
        \item \texttt{email\_address}: Hashed shopper's email (ID).
        \item \texttt{card\_number}: Hashed card number (ID).
        \item \texttt{shopper\_interaction}: Payment method (Categorical) - Ecommerce, POS. POS means an in-person or in-store transaction.
        \item \texttt{card\_bin}: Bank Identification Number (ID).
        \item \texttt{has\_fraudulent\_dispute}: Indicator of fraudulent dispute from issuing bank (Boolean).
        \item \texttt{is\_refused\_by\_adyen}: Adyen refusal indicator (Boolean).
        \item \texttt{aci}: Authorization Characteristics Indicator (Categorical).
        \item \texttt{acquirer\_country}: The location (country) of the acquiring bank (Categorical) - SE, NL, LU, IT, BE, FR, GR, ES.
    \end{itemize}
\end{itemize}

In addition to its dataset structure, DABStep categorizes tasks by difficulty and content complexity. Table~\ref{tab:dabstep-stats} summarizes key properties of the benchmark.

\begin{table}[ht]
\centering
\caption{Summary statistics of the DABStep benchmark dataset}
\label{tab:dabstep-stats}
\begin{tabular}{@{}lcc@{}}
\toprule
\textbf{Property}                     & \textbf{Value} \\ \midrule
Number of tasks                      & 456            \\
Average number of tables per task   & 3.2            \\
Median rows per table               & 5,000          \\
Percent of tasks with joins         & 64\%           \\
Percent requiring external docs     & 28\%           \\
Easy/Hard task split                 & 55\% / 45\%    \\
Eval metric                          & Exact match    \\
Domain types                         & Payments, Logistics, Inventory \\
\bottomrule
\end{tabular}
\end{table}

\subsection{Sample Tasks}

To illustrate the benchmark’s complexity and structured reasoning requirements, we present two representative examples:

\begin{itemize}
    \item \textbf{Question:} Which card scheme had the highest average fraud rate in 2023?\\
    \textbf{Guidance:} Answer must be the name of the scheme.\\
    \textbf{Answer:} \texttt{SwiftCharge}

    \vspace{0.5em}
    \item \textbf{Question:} For the year 2023, focusing on the merchant \textit{Crossfit Hanna}, if we aimed to reduce fraudulent transactions by encouraging users to switch to a different Authorization Characteristics Indicator through incentives, which option would be the most cost-effective based on the lowest possible fees?\\
    \textbf{Guidance:} Answer must be the selected ACI to incentivize and the associated cost rounded to 2 decimals in this format: \{\textit{card\_scheme}:\textit{fee}\}.\\
    \textbf{Answer:} \texttt{E:346.49}
\end{itemize}

These questions require the agent to reason over multiple dimensions: temporal filtering, column selection, aggregation, and domain knowledge interpretation.
 Although this dataset is targeted at multi-step reasoning tasks, we find it useful as it provides good documentation for the tables and the tasks, and that provides good insights.

\begin{figure}[htp]
    \centering
    \includegraphics[width=12cm]{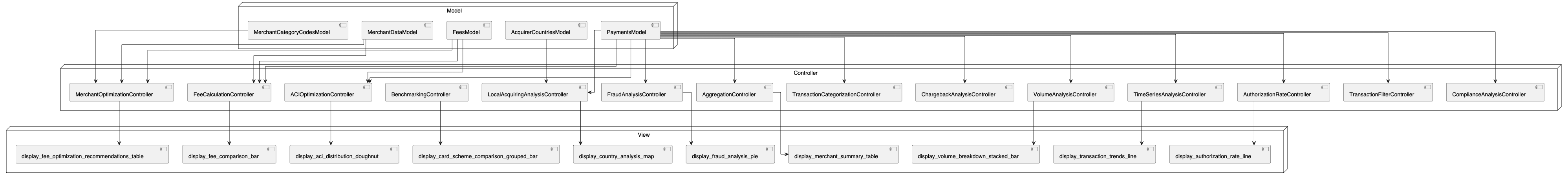}
    \caption{Model-View-Controller system for DABset expanded questions}
    \label{fig:mvc-schema}
\end{figure}

Given the schema of the database, the instructions presented in dataset DABstep, samples of the data (extracted from the data itself), and the dev-set of the dataset, we can obtain an enlarged set of questions that result in the MVC diagram presented in Figure~\ref{fig:mvc-schema}.

In this paper, we extended the work of~\cite{coelho} by instructing the generation of the MVC also to generate the interfaces to controllers and to the views, as shown in Figure~\ref{fig:mvc-schema}.

Table~\ref{table:mvc-additional-functions} presents the list of tools that were inferred from the system. Please note that we have not listed here the full interface of the functions due to lack of space. We have also instructed the LLM to utilize the information from~\cite{coolinfographics-dataviz-guides} when recommending UI artifacts in the views section. That can be seen as the suffix for the \texttt{display} functions, such as \texttt{doughnut} in \texttt{display\_aci\_distribution doughnut}.

\begin{table}
\caption{List of functions by Algorithm~\ref{fig:algorithm}, using 40 questions/query and extraction of MVC functions from MVC diagram}
\vspace*{4pt}
\label{table:mvc-additional-functions}
\small
    \centering
    \begin{tabular}{|l|}
    \hline
    \textbf{Function} \\ \hline 
    \texttt{get\_payments\_data()} \\
    \texttt{get\_fee\_rules()} \\
    \texttt{get\_merchant\_config()} \\
    \texttt{get\_acquirer\_countries()} \\
    \texttt{get\_mcc\_descriptions()} \\
    \hline
    \texttt{calculate\_transaction\_fee()} \\ %-Transaction, -MerchantConfig, -+FeeCalculationResult)} \\
    \texttt{calculate\_merchant\_fraud\_rate()} \\ %-MerchantName, -DateRange, +FraudMetrics)} \\
    \texttt{calculate\_authorization\_rate()} \\ %-MerchantName, -DateRange, +AuthMetrics)} \\
    \texttt{calculate\_chargeback\_rate()} \\ %-MerchantName, -DateRange, +ChargebackMetrics)} \\
    \texttt{categorize\_transactions()} \\ %-Transactions, -CategoryField, +CategorizedData)} \\
    \texttt{generate\_optimization\_recommendations()} \\ %-MerchantName, +Recommendations)} \\
    \texttt{analyze\_volume\_trends()} \\ %-MerchantName, -DateRange, +VolumeAnalysis)} \\
    \texttt{identify\_local\_acquiring\_opportunities()} \\ %-MerchantName, +Opportunities)} \\
    \texttt{analyze\_aci\_usage()} \\ %-MerchantName, -DateRange, +ACIAnalysis)} \\
    \texttt{check\_compliance\_metrics()} \\ %-MerchantName, +ComplianceStatus)} \\
    \texttt{filter\_transactions\_complex()} \\ %-Transactions, -ComplexFilter, +FilteredTransactions)} \\
    \texttt{aggregate\_transaction\_data()} \\ %-Transactions, -GroupByFields, -AggregateFields, +AggregatedData)} \\
    \texttt{analyze\_time\_series()} \\ %-[Header | Data], -TimeField, -ValueField, +[Header | Data])} \\
    \texttt{benchmark\_merchant\_performance()} \\ %-MerchantName, -[Header | Data], -ValueField, +[Header | Data])} \\
    \hline
    \texttt{display\_fee\_comparison\_bar()} \\
    \texttt{display\_transaction\_trends\_line()} \\
    \texttt{display\_fraud\_analysis\_pie()} \\
    \texttt{display\_merchant\_summary\_table()} \\
    \texttt{display\_aci\_distribution\_doughnut()} \\
    \texttt{display\_volume\_breakdown\_stacked\_bar()} \\
    \texttt{display\_fee\_optimization\_recommendations\_table()} \\
    \texttt{display\_authorization\_rate\_line()} \\
    \texttt{display\_country\_analysis\_map()} \\
    \texttt{display\_card\_scheme\_comparison\_grouped\_bar()} \\
    \hline
    \end{tabular}
\end{table}

\section{Conclusions}

This paper examined the evolving role of Large Language Models (LLMs) in software systems, focusing on their integration as reasoning engines and user interfaces, and the challenges this presents for software engineering, system reliability, and interpretability. We identified that while LLMs excel at probabilistic language tasks, they exhibit critical limitations when confronted with problems requiring strict logical reasoning, discrete decision-making, or adherence to formal rules. These shortcomings are particularly evident in domains that include safety-critical applications, where correctness and transparency are paramount.

To address these gaps, we proposed a neurosymbolic approach that augments LLMs with modular, composable tools—especially logic-based components such as Prolog predicates. This strategy allows LLMs to decompose complex queries into smaller, verifiable steps, orchestrating solutions by stitching together specialized, well-understood functions. By leveraging first-order logic and explicit rule systems, our methodology enables precise representation of facts, rules, and negation—capabilities that are difficult to achieve with purely probabilistic models or knowledge graphs alone.

We further demonstrated how this approach supports the development of intelligent agents, where LLMs serve as the core orchestrators, supported by structured memory, planning modules, and domain-specific tools. This architecture facilitates dynamic, context-aware reasoning and allows agents to adapt to a wide range of user queries, while maintaining reliability and interpretability. The integration of logic reasoning at the planning stage helps mitigate common failure modes of LLMs, such as hallucination and incorrect step decomposition, by providing a framework for self-reflection and evaluation of reasoning chains.

Our experiments, conducted using the DABStep benchmark, validated the effectiveness of this approach in real-world, multi-step reasoning tasks. By expanding initial user questions into comprehensive sets of sub-tasks and mapping them to Model-View-Controller (MVC) components, we were able to regain the precision and coverage previously afforded by traditional user stories and use cases. This process not only aids in system documentation and effort estimation but also supports the systematic identification of functional gaps and the design of robust, generalizable AI-driven systems.

Our work showed that combining LLMs with logic-based reasoning modules and modular tool orchestration provided a scalable path toward building AI agents capable of reliable, interpretable, and secure problem-solving across complex domains. This hybrid approach restored engineering rigor to LLM-driven systems and offers a practical methodology for both system designers and end-users seeking trustworthy AI solutions.

\section{Acknowledgments}
Portions of this document used Perplexity-AI to improve readability.  We acknowledge the valuable feedback from Luis Lamb and Ajay Kumar in reviewing earlier versions of this document.

% References
\def\refname{Reference}

\bibliography{./references.bib}
\bibliographystyle{plain}

\end{document}